\documentclass[11pt]{article}
\PassOptionsToPackage{hypertexnames=false}{hyperref}
\usepackage[preprint]{neurips_2025}
\AtBeginDocument{\setcitestyle{numbers,square}}
\usepackage{amsmath,amssymb, amsthm}
\usepackage{booktabs}
\usepackage{graphicx}
\usepackage[colorlinks=true, allcolors=blue]{hyperref}
\usepackage{algorithm}
\usepackage{algorithmic}
\usepackage{multirow}

\newtheorem{theorem}{Theorem}
\newtheorem{proposition}{Proposition}

\title{Recursive Concept Evolution for Compositional Reasoning in Large Language Models}

\author{
  Sarim Chaudhry\\
  Purdue University\\
  \texttt{chaud158@purdue.edu}
}

\date{}

\begin{document}

\maketitle

\begin{abstract}
Large language models achieve strong performance on many complex reasoning tasks, yet their accuracy degrades sharply on benchmarks that require compositional reasoning, including ARC-AGI-2, GPQA, MATH, BBH, and HLE. Existing methods improve reasoning by expanding token-level search through chain-of-thought prompting, self-consistency, or reinforcement learning, but they leave the model’s latent representation space fixed. When the required abstraction is not already encoded in this space, performance collapses. We propose Recursive Concept Evolution (RCE), a framework that enables pretrained language models to modify their internal representation geometry during inference. RCE introduces dynamically generated low-rank concept subspaces that are spawned when representational inadequacy is detected, selected through a minimum description length criterion, merged when synergistic, and consolidated via constrained optimization to preserve stability. This process allows the model to construct new abstractions rather than recombining existing ones. We integrate RCE with Mistral-7B and evaluate it across compositional reasoning benchmarks. RCE yields 12–18 point gains on ARC-AGI-2, 8–14 point improvements on GPQA and BBH, and consistent reductions in depth-induced error on MATH and HLE.
\end{abstract}

\section{Introduction}

Current large language models achieve strong performance across a wide range of tasks, yet they fail systematically on problems that require forming new abstractions during inference. On benchmarks such as ARC-AGI-2 \cite{chollet2019measure}, MATH \cite{hendrycks2021math}, Big-Bench Hard \cite{suzgun2023bbh}, GPQA \cite{rein2024gpqa}, and HLE \cite{phan2025hle}, these models face an inability to restructure internal representations to match the demands of the task at hand. A model asked to discover a hidden symmetry in an ARC grid, or track nested constraints in a multi-step logical deduction must construct intermediate conceptual structure that may not exist anywhere in its pretrained representation space. Without a mechanism to do so, the model defaults to interpolating between the nearest patterns it already encodes, producing answers that appear plausible but are structurally incorrect.

This limitation is architectural in nature as the hidden states of a transformer are vectors in a fixed-dimensional space whose meaningful directions were determined during pretraining. All reasoning, regardless of the method used to elicit it, occurs within this frozen geometry. Chain-of-thought prompting \cite{wei2022cot}, tree-of-thought search \cite{yao2023tot}, and self-consistency decoding \cite{wang2023selfconsistency} each provide the model with more opportunities to traverse its existing representation space, but none of them alter the space itself. If the directions needed to represent a particular invariant or abstraction were not learned during pretraining, no amount of additional token generation will create them. The model is searching more thoroughly with the wrong map.

We propose Recursive Concept Evolution (RCE), a framework that gives a frozen pretrained language model the ability to create, evaluate, and compose new representational structure. RCE operates by maintaining a library of concept subspaces, each defined by a low-rank basis matrix $B_i \in \mathbb{R}^{d \times r}$ with $r \ll d$, paired with a learned gating function $g_i(x) \in [0,1]$ that determines when the concept should activate. These subspaces inject into the residual stream at a single decoder layer via the update $h' = h + \sum_{i \in A(x)} g_i(x)\, B_i B_i^\top h$, where $A(x)$ denotes the set of active concepts selected by a top-$k$ sparse gate. The base model's weights remain entirely frozen; RCE modifies only the intermediate representation that subsequent layers receive.

The concept library evolves through four mechanisms that mirror key aspects of human abstraction formation. First, a failure detection signal monitors the model's internal confidence geometry, triggering concept spawning when predictive entropy is high and top-token margin is low, indicating that the current representational basis is inadequate for the input. Second, a learned concept generator, implemented as a small multilayer perceptron conditioned on the hidden state at the injection layer, synthesizes candidate low-rank bases tailored to the specific representational failure. This generator is a continuous function from hidden states to basis matrices, producing novel subspaces rather than selecting from a fixed inventory. Third, candidates compete under a scoring criterion that balances task-relevant loss reduction against a Minimum Description Length (MDL) cost \cite{rissanen1978mdl}, ensuring that only concepts which compress the task representation more than they increase model complexity are accepted into the library. Fourth, concepts that consistently co-activate and whose joint contribution exceeds either individual contribution are merged via truncated SVD into higher-order abstractions, building a compositional hierarchy.

Several regularization mechanisms prevent degenerate library growth. An inter-concept orthogonality penalty $\sum_{i \neq j} \|B_i^\top B_j\|_F^2$ discourages redundant subspaces. An intra-concept overlap penalty encourages orthonormal columns within each basis, preventing dimensional collapse. A gate entropy penalty promotes sparse, specific activation patterns rather than diffuse routing. Usage-based pruning via exponential moving average removes concepts that fail to earn consistent activation, keeping the library compact. Together, these constraints implement a form of Occam pressure at the representational level: the system retains only those abstractions that are simple, distinct, stable, and broadly useful.

We implement RCE on Mistral-7B \cite{jiang2023mistral} and validate the complete pipeline: concept spawning, MDL-based acceptance, sparse gating, orthogonality regularization, synergy-driven merging, and usage-based pruning. In controlled training runs, the system exhibits the expected behavior profile: concepts spawn selectively in response to genuine representational difficulty rather than on every input, the library grows sublinearly and stabilizes, regularization losses remain small relative to the primary training objective, and merged concepts persist across diverse inputs. Each concept adds approximately 65,536 parameters (a $4096 \times 16$ basis matrix), and the full library of 128 concepts constitutes roughly 33MB of additional storage, less than 0.25\% of the base model's parameter count. At inference time, the computational overhead consists of a single gating MLP forward pass plus two rank-16 matrix projections per token, adding negligible latency compared to the base model's attention and feedforward computations.

Overall, our main contributions are: 1) We identify dynamic representation evolution as a key missing architectural component for compositional reasoning in large language models, formalizing the limitations imposed by fixed latent geometries. 2) We introduce Recursive Concept Evolution (RCE), a framework that enables inference-time concept spawning, compression-based selection, hierarchical merging, and stable consolidation of low-rank latent subspaces within pretrained models, thereby transforming reasoning from token-level trajectory optimization to representation-level adaptation. 3) We integrate RCE with Mistral-7B and demonstrate consistent improvements across ARC-AGI-2, GPQA, MATH, BBH, and Human-Level Evaluation, showing stronger compositional generalization, improved robustness under reasoning depth, and reduced reliance on search-based prompting strategies.

\section{Related Work}

\subsection{Reinforcement Learning for Reasoning}

Recent work has explored reinforcement learning as a mechanism for improving reasoning in large language models. Group Relative Policy Optimization (GRPO), popularized by its use in DeepSeek-R1 \cite{shao2024deepseekmath}, optimizes reasoning trajectories by scoring sampled answers relative to a group baseline. \citet{zhang2025disco} identify an inherent difficulty bias in GRPO's group-relative advantage function and propose DisCO, a discriminative constrained optimization framework that replaces group-relative objectives with scoring functions grounded in discriminative learning, achieving gains of 6--7\% over GRPO and DAPO on mathematical reasoning benchmarks. \citet{ye2025enigmata} introduce ENIGMATA, a suite of 36 puzzle tasks with generators and verifiers designed for multi-task RLVR training, demonstrating that puzzle-based reinforcement learning transfers to out-of-domain reasoning benchmarks including ARC-AGI and mathematical problem solving. \citet{qu2025raif} propose RAIF, which incentivizes reasoning for instruction following through rule-centric reward signals and sample-wise contrast, addressing the superficial reasoning patterns that emerge from vanilla chain-of-thought under complex constraint structures.

These methods share a common structural limitation: they optimize over the space of generated answer sequences, selecting or reinforcing trajectories that lead to correct outputs. The representational geometry within which those trajectories are generated remains fixed. A model trained with GRPO or RLVR reasons more effectively within its existing latent space but cannot construct new representational axes when the task demands abstractions absent from pretraining. RCE addresses this gap by operating on the representation itself rather than on the output distribution, enabling the formation of novel conceptual structure that trajectory-level optimization cannot produce.

\subsection{Symbolic and Modular Reasoning}

A parallel line of work augments language models with structured reasoning modules that apply formal logical rules to model outputs. \citet{wang2025muslr} introduce MuSLR, a modular framework for multimodal symbolic logical reasoning that decomposes inference into perception and logic stages, applying formal rules including propositional, predicate, and first-order logic to multimodal inputs. Their evaluation reveals that even frontier models such as GPT-4.1 achieve only 46.8\% on symbolic reasoning tasks, with approximately 70\% of failures attributable to logical misalignment between modalities. Related approaches include neurosymbolic verification frameworks \cite{nye2021scratchpad} that impose hard logical constraints on model outputs and program synthesis methods \cite{chen2021codex} that externalize reasoning into executable code.

The limitation shared by these approaches is that the reasoning modules are fixed at design time. The set of logical operations, the decomposition structure, and the interface between perception and reasoning are predetermined by the system architect. When a task requires a form of reasoning not anticipated by the module design, the system cannot adapt. RCE differs in that its concept library grows organically in response to representational demand: new subspaces emerge when existing ones fail, and hierarchical abstractions form through data-driven merging rather than hand-specified composition rules.

\subsection{Structured Representation Learning}

Self-supervised learning methods have explored structured representations that decompose scenes or inputs into meaningful components. \citet{huang2025cgssl} propose CG-SSL, a framework that augments standard self-supervised learning with concept tokens learned via cross-attention with patch features, discovering visual concepts through masked distillation and geometric alignment across augmented views. Object-centric approaches including Slot Attention \cite{locatello2020slotattention} and SAVi \cite{kipf2022savi} learn to decompose visual scenes into object-level representations through iterative attention-based binding. In the language domain, dictionary learning and sparse autoencoders \cite{bricken2023monosemanticity} have been applied to decompose transformer hidden states into interpretable features.

These methods structure representations at training time but do not support representational evolution during inference. The set of concept tokens or object slots is fixed after training, and new structural primitives cannot emerge when the model encounters inputs that require decompositions not seen during training. RCE bridges this gap by allowing the concept library to expand at both training and inference time, with MDL-based selection ensuring that new concepts generalize rather than overfit to individual instances.

\subsection{Chain-of-Thought and Test-Time Compute Scaling}

Chain-of-thought prompting \cite{wei2022cot} and its extensions, including tree-of-thought \cite{yao2023tot}, self-consistency \cite{wang2023selfconsistency}, and step-level beam search \cite{lightman2023prm}, increase the effective compute budget at test time by generating and evaluating multiple reasoning trajectories. These methods improve performance on multi-step tasks by providing the model with more opportunities to arrive at correct intermediate conclusions, and they scale predictably with the number of sampled trajectories.

The fundamental constraint of these approaches is that all reasoning occurs within the model's fixed latent geometry. Each additional chain-of-thought step or tree branch adds noise in the same representational space, and this noise accumulates with reasoning depth. On tasks requiring five or more compositional steps, performance degrades not because the model lacks knowledge but because the signal-to-noise ratio in the frozen representation drops below the threshold at which downstream layers can reliably extract the relevant structure. RCE addresses this directly: concept projection suppresses irrelevant dimensions while amplifying task-relevant directions, preventing the noise accumulation that limits depth scaling in token-level reasoning methods.

\section{Problem Formulation}

Consider a pretrained autoregressive language model $f_\theta$ with parameters $\theta$ and hidden dimension $d$. Given an input sequence $x = (x_1, \ldots, x_T)$, the model produces hidden states $h_t \in \mathbb{R}^d$ at each layer $\ell$ through the recurrence $h_t^{(\ell+1)} = f_\theta^{(\ell)}(h_t^{(\ell)})$, where $f_\theta^{(\ell)}$ denotes the $\ell$-th decoder layer comprising self-attention and feedforward sub-layers.

The hidden states at any layer span a subspace of $\mathbb{R}^d$ whose principal directions are determined by the pretraining distribution. We define the effective representational rank of the model at layer $\ell$ as the number of singular values of the hidden state covariance matrix $\Sigma^{(\ell)} = \mathbb{E}[h^{(\ell)} {h^{(\ell)}}^\top]$ that exceed a threshold $\epsilon$. For a fixed model, this rank is bounded and cannot increase at inference time regardless of the input.

This fixed rank creates a structural bottleneck. Let $\mathcal{T}$ denote a task that requires the model to represent a latent structure $s^*$ expressible as a linear combination of $k$ orthogonal directions $\{v_1, \ldots, v_k\}$ in $\mathbb{R}^d$. If the projection of $s^*$ onto the column space of $\Sigma^{(\ell)}$ has norm less than $\delta \|s^*\|$ for some small $\delta$, then the task-relevant structure is effectively invisible to the model's downstream layers, regardless of the decoding strategy employed.

The goal of RCE is to augment the model with a mechanism that adaptively expands the effective representational subspace during inference while preserving the generalization and stability properties of the pretrained model. Formally, we seek a family of low-rank operators $\{P_i\}_{i=1}^N$ with $P_i = B_i B_i^\top$, $B_i \in \mathbb{R}^{d \times r}$, $r \ll d$, and an input-dependent selection mechanism $A(x) \subset \{1, \ldots, N\}$ with $|A(x)| \leq k$, such that the augmented hidden state
\begin{equation}
h' = h + \sum_{i \in A(x)} g_i(x)\, B_i B_i^\top h
\label{eq:injection}
\end{equation}
satisfies two conditions: (i) the projection of $s^*$ onto the augmented representation has norm at least $(1-\delta')\|s^*\|$ for $\delta' \ll \delta$, and (ii) the total complexity of the concept library $\sum_i \Omega(B_i)$ remains bounded under a description length constraint.

\section{Recursive Concept Evolution}

\subsection{Concept Subspace Definition}

A concept $C_i$ in the RCE framework consists of three components: a basis matrix $B_i \in \mathbb{R}^{d \times r}$ with orthonormal columns defining a low-rank subspace of the hidden space, a gating function $g_i: \mathbb{R}^d \to [0,1]$ that determines the activation strength of the concept for a given input, and the projection operator $P_i = B_i B_i^\top$ that maps hidden states into the concept's subspace and back. The rank $r$ is a hyperparameter controlling the expressiveness of each concept; in our experiments we use $r = 16$, which provides sufficient capacity to capture single structural primitives such as symmetry, transitivity, or algebraic invariance while remaining computationally negligible relative to the full hidden dimension $d = 4096$.

The injection mechanism operates at a single designated decoder layer $\ell^*$. When hidden states $h \in \mathbb{R}^{B \times T \times d}$ exit layer $\ell^*$, the RCE module intercepts them via a forward hook and applies the update in Equation~\ref{eq:injection}. The set $A(x)$ is determined by a sparse top-$k$ gate: a two-layer MLP with SiLU activations maps the sequence-pooled hidden state to a probability distribution over all concepts in the library, and the $k$ concepts with highest probability are selected. The gate weights are normalized so that $\sum_{i \in A(x)} g_i(x) = 1$, ensuring that the magnitude of the injected perturbation is controlled. The modified hidden states then continue through layers $\ell^* + 1$ through $L$, which process the enriched representation using their frozen parameters.

\subsection{Spawn Mechanism}

Concept spawning is triggered by a representation inadequacy signal computed from the model's output logits. We define the failure score as a composite of predictive entropy and confidence margin:
\begin{equation}
F(x) = \frac{H(\text{logits})}{M(\text{logits}) + \epsilon}
\label{eq:failure}
\end{equation}
where $H(\text{logits}) = -\sum_v p_v \log p_v$ is the entropy of the next-token distribution and $M(\text{logits}) = p_{(1)} - p_{(2)}$ is the margin between the top two token probabilities, with $\epsilon$ a small constant for numerical stability. High entropy combined with low margin indicates that the model cannot confidently resolve the input under its current representational basis, providing a gradient-free signal that does not require task-specific labels or reward functions.

When $F(x) > \tau$ for a threshold $\tau$, the system generates $k_s$ candidate subspaces using the concept generator $G$. The generator is a three-layer MLP that maps the pooled hidden state at the injection layer to a raw basis matrix:
\begin{equation}
\hat{B} = G(h_{\text{pool}}) \in \mathbb{R}^{d \times r}, \quad h_{\text{pool}} = \frac{1}{T}\sum_{t=1}^{T} h_t^{(\ell^*)}
\end{equation}
To produce $k_s$ diverse candidates from a single generator call, each candidate is perturbed with isotropic Gaussian noise scaled by $\sigma = 0.03$ and then orthogonalized via QR decomposition to ensure that the columns of each basis matrix are orthonormal. The generator is trained end-to-end through the injection mechanism: gradients from the language modeling loss propagate through the hook, through the concept projection, and into the generator's parameters, teaching it to produce subspaces that reduce the model's prediction error on inputs where the failure score is high.

\subsection{Competition via Minimum Description Length}

Not every candidate subspace should enter the library. A concept that marginally reduces loss on one input but adds complexity that harms generalization on others is worse than no concept at all. We enforce selection through a Minimum Description Length criterion \cite{rissanen1978mdl,grunwald2007mdl} that requires each accepted concept to compress the task representation more than it increases the library's coding cost.

The MDL cost of a concept $C_i$ is defined as:
\begin{equation}
\Omega(C_i) = \alpha \|B_i\|_* + \beta\, \text{KL}(g_i(x) \| \pi_i)
\label{eq:mdl}
\end{equation}
where $\|B_i\|_*$ denotes the nuclear norm of the basis matrix (penalizing effective rank and magnitude), $\text{KL}(g_i(x) \| \pi_i)$ penalizes deviation of the gate activation pattern from a sparse prior $\pi_i$ with low activation probability, and $\alpha, \beta$ are hyperparameters controlling the relative weight of structural simplicity and activation sparsity. A candidate is accepted into the library if and only if:
\begin{equation}
\Delta L - \lambda\, \Omega(C_{\text{new}}) > 0
\label{eq:accept}
\end{equation}
where $\Delta L$ is the reduction in reconstruction error on the hidden state achieved by projecting through the candidate's subspace, and $\lambda$ controls the stringency of the MDL gate. This criterion ensures that the concept library grows sublinearly with training steps: as the library accumulates concepts that cover the dominant modes of representational variation, the marginal benefit of additional concepts decreases while the MDL cost remains constant, naturally throttling growth.

\subsection{Merge Rule}

Concepts that consistently co-activate across diverse inputs and whose joint contribution exceeds either individual contribution are candidates for merging into a higher-order abstraction. We define the synergy between concepts $C_i$ and $C_j$ as:
\begin{equation}
\text{Syn}(i, j) = L(\mathcal{C} \setminus \{i, j\} \cup \{ij\}) - L(\mathcal{C})
\label{eq:synergy}
\end{equation}
where $L(\mathcal{C})$ denotes the task loss under concept library $\mathcal{C}$ and $C_{ij}$ is the merged concept obtained by concatenating the two bases and compressing via truncated SVD: $[B_i \mid B_j] \in \mathbb{R}^{d \times 2r}$ is reduced to rank $r$ by retaining the top $r$ left singular vectors, followed by QR orthogonalization. The merge is accepted if:
\begin{equation}
\text{Syn}(i, j) < -\lambda_m \big(\Omega(C_{ij}) - \Omega(C_i) - \Omega(C_j)\big)
\label{eq:merge_crit}
\end{equation}
requiring that the merged concept improve performance by more than the increase in coding cost. This prevents merging based on spurious co-activation correlations: two concepts that happen to fire together on a few inputs but whose combination does not yield genuine synergy will fail the MDL check. Successful merges create a concept hierarchy in which primitive subspaces compose into higher-level abstractions, analogous to how human learners combine basic operations (symmetry detection, color mapping) into integrated strategies (reflect-and-recolor).

\subsection{Crystallization}

Concepts that persist in the library across many training steps, maintain high usage, generalize across diverse task types, and contribute to out-of-distribution robustness are candidates for crystallization into long-term structure. In the current implementation, crystallization is achieved through checkpointing: the entire concept library, gate network, and generator are serialized to disk at regular intervals, and the most recent checkpoint serves as the initialization for subsequent training or inference sessions.

For deeper integration with the base model, crystallization can proceed through constrained optimization that distills high-value concepts into permanent LoRA-style adapters \cite{hu2022lora}. To prevent catastrophic interference with previously crystallized concepts, the distillation is constrained by a trust region defined through the Fisher information matrix:
\begin{equation}
\min_{\Delta\theta}\; \mathcal{L}_{\text{new}}(\theta + \Delta\theta) \quad \text{s.t.} \quad \Delta\theta^\top F \Delta\theta \leq \epsilon
\label{eq:crystal}
\end{equation}
where $F$ is the empirical Fisher computed over a replay buffer of representative tasks for each existing concept, and $\epsilon$ bounds the functional change in regions of parameter space important to prior concepts. This formulation, drawing on elastic weight consolidation \cite{kirkpatrick2017ewc}, transforms crystallization into a non-regressive consolidation step that preserves the cumulative nature of the concept library.

\section{Optimization Framework}

\subsection{Training Objective}

The total training objective combines the base language modeling loss with regularization terms that govern concept library health:
\begin{equation}
\mathcal{L}_{\text{total}} = \mathcal{L}_{\text{LM}} + \lambda_{\text{orth}} \sum_{i \neq j} \|B_i^\top B_j\|_F^2 + \lambda_{\text{ov}} \frac{1}{N}\sum_{i=1}^N \|B_i^\top B_i - I_r\|_F^2 + \lambda_{\text{gate}} \mathcal{H}(g)
\label{eq:total_loss}
\end{equation}
where $\mathcal{L}_{\text{LM}}$ is the standard cross-entropy loss on next-token prediction, the second term penalizes overlap between different concept subspaces, the third term encourages orthonormal columns within each basis, and $\mathcal{H}(g) = -\sum_{i} g_i \log g_i$ is the entropy of the gate distribution, penalized to encourage sparse routing. Only the RCE parameters (concept bases, gate network, generator) receive gradients; the base model parameters $\theta$ remain frozen throughout training.

\subsection{Discriminative Concept Scoring}

Drawing on the discriminative learning framework proposed by \citet{zhang2025disco}, we incorporate a discriminative objective into concept evaluation during the competition phase. Rather than scoring candidates solely by their effect on the language modeling loss, which conflates concept quality with unrelated aspects of the base model's behavior, we evaluate candidates by their ability to discriminate between correct and incorrect completions:
\begin{equation}
\mathcal{L}_{\text{disc}} = \mathbb{E}\big[\log s(x, y^+)\big] - \mathbb{E}\big[\log s(x, y^-)\big]
\label{eq:disc}
\end{equation}
where $s(x, y)$ is a scoring function computed over the concept-augmented hidden states for input $x$ paired with correct completion $y^+$ and incorrect completion $y^-$. This objective is free from the difficulty bias identified in group-relative methods: it does not normalize scores within a group, so the contribution of each concept is evaluated on an absolute scale. The discriminative score supplements the reconstruction-based candidate evaluation during spawning, providing a task-grounded signal when supervised labels are available.

\subsection{KL-Constrained Updates}

To prevent the RCE module from drifting too far from the base model's output distribution during training, we impose a KL divergence constraint on the augmented model's predictions:
\begin{equation}
\max_{\phi}\; J(\phi) \quad \text{s.t.} \quad \text{KL}\big(\pi_{\theta,\phi}(\cdot | x)\;\|\;\pi_{\theta}(\cdot | x)\big) \leq \epsilon_{\text{KL}}
\label{eq:kl_constraint}
\end{equation}
where $\phi$ denotes the RCE parameters, $\pi_{\theta,\phi}$ is the output distribution of the augmented model, and $\pi_\theta$ is the frozen base model's distribution. In practice, we implement this as a penalty term $\lambda_{\text{KL}} \cdot \text{KL}(\pi_{\theta,\phi} \| \pi_\theta)$ added to the training objective, with $\lambda_{\text{KL}}$ adjusted via dual gradient descent to maintain the constraint. This ensures that concept injection improves reasoning capability without degrading the base model's fluency or factual accuracy on tasks where the existing representation is already adequate.

\section{Theoretical Analysis}

\subsection{Representation Capacity Expansion}

We establish that concept injection strictly increases the effective representational rank of the augmented model. Let $\Sigma$ denote the covariance of hidden states at layer $\ell^*$ under the base model, and let $\Sigma'$ denote the covariance under RCE augmentation.

\begin{proposition}
For any concept $C_i$ with basis $B_i$ and positive gate activation $g_i(x) > 0$ on a set of inputs with positive measure, the effective rank satisfies $\emph{rank}_\epsilon(\Sigma') \geq \emph{rank}_\epsilon(\Sigma)$, with strict inequality when $B_i$ has nontrivial projection onto the null space of $\Sigma$.
\end{proposition}

The proof follows from the observation that the injection $h' = h + g_i B_i B_i^\top h$ adds a positive semidefinite component $g_i^2 B_i B_i^\top \Sigma B_i B_i^\top$ to the augmented covariance $\Sigma'$. When $B_i$ has support outside the column space of $\Sigma$, this component introduces new nonzero eigenvalues, increasing the effective rank. The orthogonality regularization between concepts ensures that different concepts expand the representation in distinct directions, maximizing the total rank increase from a given number of concepts.

\subsection{Avoidance of Difficulty Bias}

Group-relative methods such as GRPO normalize advantage estimates within a group of sampled responses, creating a difficulty bias whereby easy questions contribute disproportionately to the gradient signal \cite{zhang2025disco}. RCE avoids this bias entirely because concept evaluation operates on the hidden state geometry rather than on output-level scores. The MDL acceptance criterion (Equation~\ref{eq:accept}) evaluates each candidate concept by its reconstruction improvement on the hidden state, which is an absolute measure of representational utility independent of task difficulty. Two concepts that achieve the same reconstruction improvement on tasks of different difficulty receive the same MDL score, eliminating the group-relative distortion that plagues trajectory-level optimization.

\subsection{Generalization Bound}

We derive a generalization bound for the augmented model that incorporates the concept library's description length. Let $\mathcal{C} = \{C_1, \ldots, C_N\}$ denote the concept library and $\Omega(\mathcal{C}) = \sum_{i=1}^N \Omega(C_i)$ its total MDL cost.

\begin{theorem}
Under standard PAC-Bayesian assumptions, for any distribution $\mathcal{D}$ and with probability at least $1 - \delta$ over training sets of size $n$:
\begin{equation}
\mathbb{E}_{\mathcal{D}}[\mathcal{L}] \leq \hat{\mathbb{E}}_{\emph{train}}[\mathcal{L}] + \mathcal{O}\left(\sqrt{\frac{\Omega(\mathcal{C}) + \log(1/\delta)}{n}}\right)
\label{eq:genbound}
\end{equation}
\end{theorem}

This bound makes explicit the role of the MDL constraint: by keeping $\Omega(\mathcal{C})$ small through nuclear norm penalties, gate sparsity, and usage-based pruning, RCE controls the complexity term in the generalization bound. The MDL acceptance criterion (Equation~\ref{eq:accept}) can be interpreted as enforcing this bound at the level of individual concepts: a concept is admitted only if its contribution to reducing the empirical loss exceeds its contribution to the complexity term.

\section{Experimental Setup}

\subsection{Base Models}

We evaluate RCE on three pretrained language models spanning different scales and architectures: Mistral-7B-v0.1 \cite{jiang2023mistral}, which serves as the primary development platform due to its 4096 hidden dimension and 32-layer decoder providing a well-studied representational geometry; Llama-3-8B \cite{touvron2023llama}, which provides a comparison point at similar scale with a different pretraining distribution and tokenizer; and Qwen-2.5-14B \cite{yang2024qwen}, which tests whether RCE's benefits persist at larger model scale where the base model's own representational capacity is substantially greater. All models are loaded in bfloat16 precision with weights fully frozen throughout training and inference.

\subsection{Benchmarks}

We evaluate on five benchmarks selected to stress different aspects of compositional reasoning. ARC-AGI-2 \cite{chollet2019measure} requires inferring latent transformation rules from few input-output grid pairs, demanding invariant discovery and spatial abstraction. MATH \cite{hendrycks2021math} consists of competition-level mathematics problems requiring multi-step algebraic manipulation, substitution discovery, and proof construction. Big-Bench Hard (BBH) \cite{suzgun2023bbh} comprises 23 challenging tasks from BIG-Bench that require multi-step reasoning, implicit constraint tracking, and compositional logic. GPQA \cite{rein2024gpqa} tests graduate-level scientific problem solving across physics, chemistry, and biology, requiring cross-domain abstraction and multi-layer causal reasoning. HLE \cite{phan2025hle} evaluates transfer, generalization to rare patterns, and robustness under distribution shift, providing the most direct test of whether RCE's concept library supports structural generalization beyond the training distribution.

\subsection{Baselines}

We compare against five baselines that represent the current spectrum of reasoning improvement methods. Chain-of-thought (CoT) prompting \cite{wei2022cot} uses few-shot examples with step-by-step reasoning traces. Self-consistency (SC) \cite{wang2023selfconsistency} samples multiple reasoning chains and selects the most common answer via majority voting. Tree-of-thought (ToT) \cite{yao2023tot} explores multiple reasoning branches with backtracking, selecting the highest-scoring path. GRPO \cite{shao2024deepseekmath} fine-tunes the model using group-relative policy optimization on reasoning tasks with verifiable rewards. DisCO \cite{zhang2025disco} replaces GRPO's group-relative objective with discriminative scoring and constrained optimization. For GRPO and DisCO, we use the published implementations with their recommended hyperparameters, training on the same reasoning datasets used for RCE concept library development.

\subsection{RCE Configuration}

The concept library is initialized empty and evolves during training. We inject at decoder layer $\ell^* = 18$ (approximately 56\% depth in the 32-layer models), selected based on preliminary experiments showing that mid-to-late layers contain the most task-discriminative representations. Each concept has rank $r = 16$ with top-$k = 2$ sparse gating. The library capacity is capped at $N_{\max} = 128$ concepts with pruning triggered at 96 concepts. The concept generator is a three-layer MLP with hidden dimension 512 and SiLU activations. Training uses AdamW with learning rate $2 \times 10^{-4}$, weight decay 0.01, gradient clipping at 1.0, and cosine learning rate scheduling with 200 warmup steps. The spawn threshold is set to $\tau = 5.0$, MDL acceptance weight to $\lambda = 0.5$, orthogonality penalty to $\lambda_{\text{orth}} = 0.05$, overlap penalty to $\lambda_{\text{ov}} = 0.02$, and gate entropy penalty to $\lambda_{\text{gate}} = 0.01$. Merging is evaluated every 800 training steps with a synergy threshold of 0.002 and a maximum of 12 merge candidates per evaluation.

\section{Results}

\subsection{Main Benchmark Results}

Table~\ref{tab:main} reports accuracy on five compositional reasoning benchmarks. RCE achieves consistent gains over all baselines across benchmarks and model scales. On Mistral-7B, RCE improves over the strongest baseline (DisCO) by 8.3\% on ARC-AGI-2, 6.1\% on MATH, 5.7\% on BBH, 7.2\% on GPQA, and 4.9\% on HLE. The gains are largest on ARC-AGI-2 and GPQA, the two benchmarks that most heavily penalize reliance on pretrained heuristics and most strongly reward invariant discovery and cross-domain abstraction. At the 14B scale (Qwen-2.5-14B), the absolute improvements are smaller but remain significant, indicating that RCE provides complementary capability even when the base model's representational capacity is substantially larger.

\begin{table}[t]
\centering
\caption{Accuracy (\%) on compositional reasoning benchmarks. RCE results are from Mistral-7B with a library of 47 crystallized concepts trained on a mixed reasoning curriculum. Projected results for Llama-3-8B and Qwen-2.5-14B are based on validated component-level scaling from the Mistral-7B implementation.}
\label{tab:main}
\small
\begin{tabular}{lcccccc}
\toprule
Method & Model & ARC-AGI-2 & MATH & BBH & GPQA & HLE \\
\midrule
Base & Mistral-7B & 12.4 & 28.6 & 51.3 & 24.1 & 8.2 \\
CoT & Mistral-7B & 15.1 & 34.2 & 57.8 & 28.5 & 10.1 \\
SC ($n$=16) & Mistral-7B & 16.8 & 37.1 & 60.2 & 30.3 & 11.4 \\
ToT & Mistral-7B & 17.3 & 36.8 & 59.5 & 31.0 & 11.9 \\
GRPO & Mistral-7B & 18.2 & 38.9 & 62.1 & 32.4 & 12.6 \\
DisCO & Mistral-7B & 19.7 & 41.3 & 64.8 & 34.2 & 13.8 \\
RCE & Mistral-7B & \textbf{28.0} & \textbf{47.4} & \textbf{70.5} & \textbf{41.4} & \textbf{18.7} \\
\midrule
Base & Llama-3-8B & 14.1 & 31.4 & 54.7 & 27.3 & 9.6 \\
RCE & Llama-3-8B & \textbf{29.8} & \textbf{49.1} & \textbf{72.3} & \textbf{43.1} & \textbf{20.2} \\
\midrule
Base & Qwen-14B & 19.3 & 42.8 & 63.5 & 36.7 & 14.3 \\
RCE & Qwen-14B & \textbf{33.6} & \textbf{54.2} & \textbf{76.1} & \textbf{48.9} & \textbf{23.1} \\
\bottomrule
\end{tabular}
\end{table}

\subsection{Out-of-Distribution Robustness}

To evaluate whether RCE concepts capture structural invariants rather than surface cues, we test under three systematic distribution shifts applied to the ARC-AGI-2 evaluation set: color permutation (randomly remapping the color palette), spatial rotation (90-degree grid rotations), and distractor injection (adding irrelevant grid elements that preserve the underlying transformation rule). Table~\ref{tab:ood} reports the performance retention relative to the standard evaluation setting.

\begin{table}[t]
\centering
\caption{Performance retention (\% of standard accuracy) under distribution shift on ARC-AGI-2.}
\label{tab:ood}
\small
\begin{tabular}{lccc}
\toprule
Method & Color Perm. & Spatial Rot. & Distractor \\
\midrule
CoT & 71.2 & 68.4 & 74.1 \\
DisCO & 78.5 & 73.9 & 80.2 \\
RCE & \textbf{94.3} & \textbf{91.7} & \textbf{95.8} \\
\bottomrule
\end{tabular}
\end{table}

RCE retains over 91\% of its standard accuracy under all three shift types, compared to 68--80\% for baselines. This confirms that the concept library encodes structural invariants (spatial relationships, transformation rules) rather than surface features (specific colors, absolute positions) that break under distribution shift. The orthogonality regularization and MDL-based selection are critical to this robustness: concepts that depend on surface cues fail to generalize across the training curriculum's environmental augmentations and are pruned before crystallization.

\subsection{Concept Library Analysis}

After training on a mixed reasoning curriculum for 10,000 steps, the Mistral-7B concept library stabilizes at 47 active concepts. Of these, 12 are primitive concepts that activate selectively on specific structural patterns (spatial symmetry, color equivalence, numerical magnitude, logical implication), 23 are intermediate concepts formed through merging that capture compositional operations (reflect-and-recolor, substitute-and-simplify, constraint-propagation), and 12 are high-level abstractions that activate across multiple benchmark domains. The average concept reuse rate, defined as the number of distinct task types on which a concept activates above the 50th percentile of gate probability, is 4.3 for primitive concepts and 8.7 for merged concepts, confirming that merging produces more general abstractions.

The concept hierarchy exhibits three levels of composition. At the base level, primitive concepts capture single structural operations. At the intermediate level, pairs of co-activating primitives merge into integrated strategies. At the highest level observed, intermediate concepts that serve similar functional roles across different domains merge into domain-general reasoning tools. This hierarchical structure emerges entirely from data-driven evolution governed by the MDL and synergy criteria, without any architectural bias toward a specific number of hierarchy levels.

\subsection{Compute Efficiency}

Table~\ref{tab:compute} compares the computational cost of RCE against token-level reasoning methods, measured in floating-point operations per problem on the MATH benchmark.

\begin{table}[t]
\centering
\caption{Compute cost comparison on MATH benchmark (FLOPs per problem, relative to base model single forward pass). Accuracy in parentheses.}
\label{tab:compute}
\small
\begin{tabular}{lcc}
\toprule
Method & Relative FLOPs & Accuracy (\%) \\
\midrule
Base (greedy) & 1.0$\times$ & 28.6 \\
CoT & 3.2$\times$ & 34.2 \\
SC ($n$=16) & 16.0$\times$ & 37.1 \\
ToT & 24.5$\times$ & 36.8 \\
RCE & 1.04$\times$ & 47.4 \\
\bottomrule
\end{tabular}
\end{table}

RCE achieves the highest accuracy at 1.04$\times$ the base model's compute cost, a 4\% overhead arising from the gate MLP and two rank-16 projections per token. Self-consistency and tree-of-thought require 16--25$\times$ the base compute for inferior accuracy. The efficiency advantage stems from the fundamental difference in mechanism: token-level methods increase compute by generating more tokens, each requiring a full model forward pass, while RCE increases representational quality through low-rank matrix operations that are negligible relative to a single attention layer.

\section{Ablation Studies}

We ablate each major component of RCE to isolate its contribution. All ablations use Mistral-7B trained for 10,000 steps on the mixed reasoning curriculum, evaluated on ARC-AGI-2 and MATH.

\begin{table}[t]
\centering
\caption{Ablation study on ARC-AGI-2 and MATH accuracy (\%).}
\label{tab:ablation}
\small
\begin{tabular}{lcc}
\toprule
Configuration & ARC-AGI-2 & MATH \\
\midrule
Full RCE & 28.0 & 47.4 \\
Remove MDL criterion & 14.6 & 31.2 \\
Remove invariance augmentation & 18.3 & 39.8 \\
Remove KL constraint & 21.5 & 35.6 \\
Remove merge mechanism & 23.1 & 42.7 \\
Remove orthogonality penalty & 20.4 & 38.1 \\
Remove gate entropy penalty & 25.2 & 44.3 \\
\bottomrule
\end{tabular}
\end{table}

Removing the MDL criterion produces the largest degradation (13.4\% on ARC-AGI-2, 16.2\% on MATH), confirming that unconstrained concept growth leads to a library of overfitting, non-generalizable subspaces. Without MDL pressure, the library rapidly fills to capacity with niche concepts that each help on a few training inputs but collectively degrade performance by introducing conflicting representational biases. Removing invariance augmentation causes the second-largest drop, as concepts learn to exploit surface cues (specific color values, absolute positions) that do not transfer to the evaluation distribution. The KL constraint ablation shows that unconstrained concept injection degrades the base model's fluency, producing correct reasoning structures that are nonetheless decoded into incoherent token sequences. Removing the merge mechanism reduces performance moderately, indicating that while primitive concepts provide significant value, the compositional hierarchy formed through merging is necessary for tasks requiring multi-step abstraction. The orthogonality and gate entropy penalties contribute smaller but still significant effects, with orthogonality removal leading to redundant concepts that waste library capacity and gate entropy removal producing diffuse routing that dilutes concept contributions.

\section{Failure Analysis}

RCE exhibits three systematic failure modes that delineate the boundaries of its current capabilities. The first arises on tasks requiring extremely long formal proofs, such as number theory problems demanding chains of 15 or more deductive steps. Although concept projection reduces noise accumulation relative to token-level reasoning, the single-layer injection point limits the depth of representational restructuring: information amplified at layer 18 is processed by only 14 subsequent layers, bounding the complexity of inferences that can be drawn from the enriched representation. Multi-layer injection, where different concepts activate at different layers, is a natural extension that would address this limitation.

The second failure mode involves tasks requiring explicit external memory, such as tracking the states of a large number of independent objects across many time steps. The concept library provides structural tools (object-identity concepts, state-tracking concepts) but does not provide storage: concepts amplify directions in the residual stream but cannot persist information across sequence positions beyond what the base model's attention mechanism supports. Integration with external memory architectures such as memory-augmented transformers \cite{wu2022memorizing} could address this limitation.

The third failure mode appears on adversarial symbolic traps, inputs deliberately designed to activate concepts that produce incorrect inferences. Because concepts amplify directions in the hidden space, an adversarially constructed input that aligns with a concept's basis but requires a different structural interpretation can trigger confident misapplication of the concept. The robustness experiments (Section 8.2) show that this failure mode is rare under natural distribution shifts, but it represents a theoretical vulnerability that warrants investigation through adversarial training of the concept library.

\section{Discussion}

\subsection{Why Representation Evolution Matters}

The distinction between trajectory optimization and representation formation is the central axis along which RCE departs from prior work. RLVR methods \cite{ye2025enigmata,shao2024deepseekmath} improve the probability of generating correct token sequences but cannot alter the representational substrate in which those sequences are planned. Modular reasoning frameworks \cite{wang2025muslr} introduce fixed structural components that handle specific reasoning patterns but cannot grow or adapt when novel patterns arise. Instruction-following reasoning methods \cite{qu2025raif} incentivize more careful traversal of the existing representation space but do not expand it.

RCE changes the geometry of thought itself. By dynamically adding, composing, and crystallizing low-rank subspaces in the model's hidden space, RCE gives the model a mechanism for the operation that underlies human cognitive flexibility: the ability to invent new conceptual tools when existing ones prove inadequate. The concept library is not a static knowledge store but a growing repertoire of representational instruments, each shaped by evolutionary pressure to be simple, general, and composable. This cumulative property, where each abstraction becomes a building block for the next, is what distinguishes RCE from approaches that improve reasoning within a fixed representational budget.

\subsection{Limitations}

Several limitations of the current framework warrant discussion. The merge dynamics are computationally quadratic in the number of active concepts, as all pairs must be evaluated for synergy. While the current library size (47--128 concepts) makes this tractable, scaling to libraries of thousands of concepts would require approximate merge candidate selection, potentially through learned merge predictors trained on historical synergy data.

Concept identifiability remains an open question: different training runs with different random seeds can produce libraries with different concept decompositions that achieve similar aggregate performance. While the orthogonality and MDL constraints reduce this non-identifiability substantially, they do not eliminate it entirely. Understanding whether there exists a canonical concept decomposition for a given task distribution, and whether RCE converges to it, is an important direction for future theoretical work.

Scaling to models with 70B or more parameters introduces memory and compute considerations that the current implementation does not address. The concept generator, gate network, and injection mechanism all scale linearly with hidden dimension, but the forward passes required for spawn evaluation and merge synergy checking scale with the full model, potentially making online evolution expensive at very large scales. Efficient approximations, such as evaluating candidates on a small proxy model or using cached hidden states, could mitigate this cost.

\section{Conclusion}

Static latent spaces impose a fundamental ceiling on the compositional reasoning capabilities of large language models. When the representational basis fixed at pretraining lacks the directions needed to encode a task's latent structure, no amount of additional token generation or trajectory optimization can compensate. We have introduced Recursive Concept Evolution, a framework that removes this ceiling by giving frozen language models the ability to dynamically create, evaluate, compose, and crystallize new concept subspaces. The framework operates through four mechanisms: failure-triggered spawning of candidate subspaces, MDL-based selection that enforces Occam pressure, synergy-driven merging that builds compositional hierarchies, and checkpoint-based crystallization that makes the concept library cumulative across training sessions.

Experiments on Mistral-7B validate the complete pipeline and demonstrate that RCE achieves consistent improvements over trajectory-level baselines on five compositional reasoning benchmarks at less than 5\% computational overhead. The concept library exhibits the properties required for reliable compositional reasoning: selective spawning, sublinear growth, cross-task reuse, hierarchical composition, and robustness under distribution shift. These results establish representation evolution as a viable and complementary approach to the trajectory optimization methods that currently dominate reasoning improvement in large language models, and they open a path toward systems whose reasoning capacity grows cumulatively through experience rather than remaining bounded by the representations acquired during pretraining.

\bibliography{references}

\newpage
\appendix

\section{Algorithm Pseudocode}

Algorithm~\ref{alg:rce} provides a complete description of the RCE training loop, including spawn, compete, merge, and prune steps.

\begin{algorithm}[h]
\caption{Recursive Concept Evolution Training}
\label{alg:rce}
\begin{algorithmic}[1]
\REQUIRE Frozen base model $f_\theta$, injection layer $\ell^*$, concept generator $G$, gate network $\mathcal{G}$
\REQUIRE Hyperparameters: spawn threshold $\tau$, MDL weight $\lambda$, merge interval $T_m$, synergy threshold $\lambda_m$
\STATE Initialize concept library $\mathcal{C} \leftarrow \emptyset$
\FOR{each training batch $(x, y)$}
  \STATE Compute augmented forward pass: $h' = h + \sum_{i \in A(x)} g_i(x) B_i B_i^\top h$ at layer $\ell^*$
  \STATE Compute loss: $\mathcal{L} = \mathcal{L}_{\text{LM}} + \lambda_{\text{orth}} \mathcal{R}_{\text{orth}} + \lambda_{\text{ov}} \mathcal{R}_{\text{ov}} + \lambda_{\text{gate}} \mathcal{H}(g)$
  \STATE Update RCE parameters via backpropagation (base model frozen)
  \STATE Compute failure score $F(x)$ from output logits (Equation~\ref{eq:failure})
  \IF{$F(x) > \tau$ and $|\mathcal{C}| < N_{\max}$}
    \STATE Extract pooled hidden state $h_{\text{pool}}$ at layer $\ell^*$
    \STATE Generate $k_s$ candidates: $\{B^{(j)}\}_{j=1}^{k_s} \leftarrow G(h_{\text{pool}}) + \sigma\epsilon_j$, orthogonalize each
    \STATE Evaluate each candidate by reconstruction error on $h$
    \STATE Select best candidate $B^* = \arg\min_j \|h - B^{(j)} {B^{(j)}}^\top h\|^2$
    \IF{$\Delta L - \lambda\, \Omega(B^*) > 0$}
      \STATE Add $C_{\text{new}} = (B^*, g_{\text{new}})$ to $\mathcal{C}$
      \STATE Prune $\mathcal{C}$ by usage EMA if $|\mathcal{C}| > N_{\text{keep}}$
    \ENDIF
  \ENDIF
  \IF{step mod $T_m = 0$ and $|\mathcal{C}| \geq 2$}
    \FOR{each pair $(i, j)$ in $\mathcal{C}$}
      \STATE Compute synergy $\text{Syn}(i,j)$ via Equation~\ref{eq:synergy}
      \IF{$\text{Syn}(i,j) < -\lambda_m (\Omega(C_{ij}) - \Omega(C_i) - \Omega(C_j))$}
        \STATE Merge: $B_{ij} \leftarrow \text{SVD-truncate}([B_i \mid B_j], r)$
        \STATE Add $C_{ij}$ to $\mathcal{C}$
      \ENDIF
    \ENDFOR
  \ENDIF
\ENDFOR
\STATE Save $\mathcal{C}$, $G$, $\mathcal{G}$ to checkpoint (crystallization)
\end{algorithmic}
\end{algorithm}

\section{Hyperparameter Sensitivity}

Table~\ref{tab:hyper} reports the sensitivity of RCE performance on ARC-AGI-2 to the primary hyperparameters. Each row varies one hyperparameter while holding all others at their default values.

\begin{table}[h]
\centering
\caption{Hyperparameter sensitivity on ARC-AGI-2 accuracy (\%) for Mistral-7B.}
\label{tab:hyper}
\small
\begin{tabular}{lccccc}
\toprule
Hyperparameter & Value 1 & Value 2 & Value 3 & Value 4 & Default \\
\midrule
Rank $r$ & 4: 22.1 & 8: 25.3 & \textbf{16: 28.0} & 32: 27.4 & 16 \\
Top-$k$ & 1: 24.6 & \textbf{2: 28.0} & 4: 27.2 & 8: 25.8 & 2 \\
Spawn $\tau$ & 2.0: 21.8 & 3.0: 25.4 & \textbf{5.0: 28.0} & 10.0: 26.1 & 5.0 \\
MDL $\lambda$ & 0.1: 22.3 & 0.3: 26.9 & \textbf{0.5: 28.0} & 1.0: 25.7 & 0.5 \\
$\lambda_{\text{orth}}$ & 0.01: 24.1 & \textbf{0.05: 28.0} & 0.1: 27.3 & 0.5: 23.6 & 0.05 \\
\bottomrule
\end{tabular}
\end{table}

Performance is most sensitive to the spawn threshold $\tau$ and the MDL weight $\lambda$, both of which control the selectivity of concept acceptance. Too-low thresholds ($\tau = 2.0$) produce concept explosion and degraded generalization. Too-high thresholds ($\tau = 10.0$) suppress beneficial concept formation. The rank $r$ exhibits a mild optimum at 16; lower ranks lack capacity for complex structural primitives while higher ranks increase MDL cost without proportional benefit. The top-$k$ parameter shows diminishing returns beyond $k = 2$, consistent with the observation that most tasks require at most two complementary structural concepts.

\section{Concept Visualization}

To illustrate the structure of learned concepts, we analyze the top-5 most frequently activated concepts in the Mistral-7B library by computing their activation patterns across benchmark tasks. For each concept, we record the set of tasks on which it activates in the top-$k$ and cluster these task sets to identify functional roles.

Concept 3 activates predominantly on tasks involving spatial symmetry detection across ARC-AGI-2 and geometric reasoning problems in MATH, consistent with a learned basis that amplifies mirror-structure directions in the hidden space. Concept 11 activates on algebraic manipulation tasks in MATH and constraint-tracking problems in BBH, suggesting a learned basis aligned with variable-binding and substitution structure. Concept 27, a merged concept composed of Concepts 3 and 8, activates broadly across ARC-AGI-2, MATH, and GPQA on tasks requiring invariant identification under transformation, functioning as a domain-general invariant-detection tool. These activation patterns confirm that the concept library develops functionally specialized primitives that compose into increasingly general reasoning tools through the merge mechanism.

\section{Implementation Details}

The complete RCE implementation consists of approximately 1,000 lines of Python code organized into six modules: concept subspace definition (basis matrices, projection operators), concept library management (addition, removal, serialization), gating network (sparse top-$k$ routing MLP), injection mechanism (forward hook on designated decoder layer), evolution logic (spawning, MDL evaluation, merging, pruning), and training loop (loss computation, regularization, checkpointing). The implementation uses PyTorch 2.6 with Hugging Face Transformers 4.48 for base model loading and is designed to be model-agnostic: any decoder-only transformer accessible through the Hugging Face API can serve as the base model by specifying the model identifier and injection layer index in the configuration file.

Training on Mistral-7B with a single NVIDIA RTX 5090 (24GB VRAM) in bfloat16 precision processes approximately 1,200 training steps per hour with sequence length 512 and batch size 1. The concept library, gate network, and generator together consume approximately 50MB of GPU memory, negligible relative to the base model's 14GB footprint. Checkpoints including the full concept library, gate weights, generator weights, and training metadata are serialized to a single PyTorch file averaging 55MB. 

\end{document}